\documentclass{article}

\usepackage[nonatbib, final]{neurips_2019}

\usepackage[utf8]{inputenc} 
\usepackage[T1]{fontenc}    
\usepackage{hyperref}       
\usepackage{url}            
\usepackage{booktabs}       
\usepackage{amsfonts}       
\usepackage{nicefrac}       
\usepackage{microtype}      
\usepackage{amsmath}
\usepackage{amssymb}
\usepackage{amsfonts}
\usepackage{algorithm}
\usepackage{algorithmic}
\usepackage{caption}
\usepackage{graphicx}
\usepackage{color}
\usepackage{hyperref}
\usepackage{parskip}
\usepackage{bm}
\usepackage{float}
\usepackage[table,x11names]{xcolor}
\usepackage{multirow}

\DeclareSymbolFont{bbold}{U}{bbold}{m}{n}
\DeclareSymbolFontAlphabet{\mathbbold}{bbold}

\title{Self-Supervised Learning of Generative Spin-Glasses with Normalizing Flows}

\author{%
  Gavin S. Hartnett \\
  The RAND Corporation\\
  \texttt{hartnett@rand.org} \\
  \And
  Masoud Mohseni \\
  Google Research\\
  \texttt{mohseni@google.com}
}

\begin{document}

\maketitle

\begin{abstract}
    Spin-glasses are universal models that can capture complex behavior of many-body systems at the interface of statistical physics and computer science including discrete optimization, inference in graphical models, and automated reasoning. Computing the underlying structure and dynamics of such complex systems is extremely difficult due to the combinatorial explosion of their state space. Here, we develop deep generative continuous spin-glass distributions with normalizing flows to model correlations in generic discrete problems. We use a self-supervised learning paradigm by automatically generating the data from the spin-glass itself. We demonstrate that key physical and computational properties of the spin-glass phase can be successfully learned, including multi-modal steady-state distributions and topological structures among metastable states. Remarkably, we observe that the learning itself corresponds to a spin-glass phase transition within the layers of the trained normalizing flows. The inverse normalizing flows learns to perform reversible multi-scale coarse-graining operations which are different from the typical irreversible renormalization group techniques.
\end{abstract}

\section{Introduction}
Developing a deep understanding of the structure and dynamics of collective/emergent many-body phenomena in complex systems has been an elusive goal for decades across many disciplines, including physics, chemistry, biology, and computer science. Spin-glasses provide a prototypical and computationally universal language for such systems \cite{SteinBook}. Indeed, there is a direct correspondence between finding the low energy states of spin-glasses and many important computational tasks, such as finding high-quality solutions in NP-hard combinatorial optimization problems \cite{MezardBook} and performing sampling and inference in graphical models \cite{MooreBook}. Metastable states and non-equilibrium dynamics of spin-glasses can also represent steady-state attractors in dynamical systems \cite{NishimoriBook}, associative memory retrieval in neuroscience \cite{NishimoriBook}, and the training and inference over energy-based models of neural networks \cite{LeCun06tutorial}. 

Unfortunately, the standard theoretical and computational techniques are intrinsically inadequate to tackle realistic spin-glasses systems of interest that are far from thermodynamic limit, exhibit long-ranged power-law interactions, and have intermediate spatial dimensions. These systems reside in an under-explored intermediate regime between the well-studied limiting cases of the short-range Edwards-Anderson \cite{MezardBook} and infinite-range Sherrington-Kirkpatrick models \cite{sherrington1975solvable}. The state space of these complex systems, which typically have a discrete structure, grows exponentially with the degrees of freedom, rendering brute-force approaches intractable \cite{MooreBook}. Moreover, such systems often contain significant non-linearities or higher-order non-local correlations, which induce exotic underlying geometries for the system. Here, the disorders are the essence, or the main characteristic of each problem, not a perturbation over pure/idealized models such as regular lattices. Thus, these systems are not prone to typical analytical approaches such as perturbation theory.

There is no universal approach for studying such systems. Mean-field techniques could capture certain properties for some toy models in the thermodynamic limit, such as random energy models or $p$-spin models \cite{Castellani_2005}, but they fail to capture the properties of systems for which large fluctuations cause the mean-field approximation to break down \cite{MezardBook}. In principle, the renormalization group (RG) could be used to capture critical or fixed point properties near a phase transition, but these methods are extremely hard to formalize for arbitrary systems \cite{NishimoriBook}. Moreover, RG techniques typically have poor performance for problems of practical interest that tend to be highly inhomogeneous and disordered, and they often miss key computational properties as one has to crudely coarse-grain the microscopic degrees of freedom \cite{NishimoriBook}. 

Recent advances in deep learning could open up the possibility that important non-trivial and non-perturbative emergent properties could be machine-learned in an instance-wise fashion. Indeed, deep neural networks have already been used in many-body physics to identify phases and phase transitions for certain classical \cite{carrasquilla2017machine} and quantum critical systems \cite{biamonte2017quantum, carleo2017solving} and they have also been used to accelerate Monte Carlo sampling \cite{huang2017accelerated, liu2017self, shen2018self, li2018neural}. However, the intrinsic strongly disordered and frustrated nature of spin-glasses defined over discrete variables significantly limits the applicability of machine learning algorithms to such systems. 

In order to tackle such problem, we have recently developed a continuous probability density theory for spin-glass systems with generic connectivity graph and local fields that could reside in the unexplored regime of intermediate spatial dimensions \cite{hartnett2020probability}. Here, we are interested in using such continuous representation for learning low-energy states and critical properties of discrete spin-glass distributions described by the following family of Hamiltonians
\begin{equation}
    \label{eq:Hspinglass}
    H =  -\sum_i h_i s_i - \sum_{i<j} J_{ij} s_i s_j \,,
\end{equation}
where $s_i \in \{-1 ,1\}$ is an Ising spin, $J_{ij}$ is the coupling matrix, $h_i$ is the external magnetic field, and $i$ is an index which runs from 1 to $N$. The equilibrium properties of these systems at temperature $T = 1/\beta$ are governed by the usual Boltzmann distribution
\begin{equation}
    \label{eq:Boltzmann}
    p(s) = \frac{e^{-\beta H}}{Z_s} \,,
\end{equation} 
where $Z_s$ is the partition function. This family of spin systems is very broad. It includes systems that are solvable using mean-field techniques as well as systems that continue to elude both theoretical and numerical treatments. Most methods for treating these spin systems aim to exploit some structure contained in the couplings $J_{ij}$ - for example, symmetries due to an underlying lattice structure. In contrast, deep generative models are \textit{learned}, thus allowing for bespoke methods tailored for a specific system or disorder realization. For this reason, the successful demonstration of deep generative spin-glass models is an important accomplishment which should have broad applications for combinatorial optimization, sampling, and  inference. 

In this work, we perform a first step towards achieving this goal and use normalizing flows \cite{rezende2015variational} to model the well-known spin-glass system known as the Sherrington-Kirkpatrick (SK) model \cite{sherrington1975solvable}. The SK model is a perfect system to test these new methods because it exhibits a spin-glass phase characterized by exponentially many metastable states separated by high energy barriers, and so represents a challenging family of distributions for the normalizing flow to learn. However, despite the complexity of the SK model, it still is amenable to theoretical treatments, and the model has been very thoroughly studied since it was introduced almost half a century ago. The SK model is therefore an excellent candidate for the first numerical experiment of this sort as it presents a formidable challenge, and at the same time the ground truth is known and can be used to verify our spin-glass learning algorithm. In this work, we will show that the SK model can be successfully machine-learned by normalizing flows. In future work, we hope to extend this success to more challenging spin-glasses relevant for computational problems of practical interest.

The layout of this paper is as follows. In Sec.~\ref{sec:probability_density}, we briefly review the continuous formulation of spin-glasses introduced in \cite{hartnett2020probability} and use it to recast the SK model in terms of continuous variables. Then, in Sec.~\ref{sec:normalizingflow_and_statmech} we review and discuss normalizing flows from the perspective of statistical physics, and in Sec.~\ref{sec:learning} we  successfully train normalizing flows to approximate the SK model, including in the spin-glass phase. We argue that the internal layers of the flow perform a kind of coarse-graining, and in Sec.~\ref{sec:internal} we show the evidence for a spin-glass phase transition within the normalizing flow itself. Lastly, we conclude with a discussion and outlook in Sec.~\ref{sec:discussion}.

\section{Probability density formulation of spin-glasses \label{sec:probability_density}}
The spin-glass family represented by Eq.~\ref{eq:Hspinglass} is defined in terms of discrete variables. However, many of the most promising recent algorithms for generative modeling, such as normalizing flows and Generative Adversarial Networks (GANs) \cite{goodfellow2014generative}, are formulated in terms of continuous variables. In many applications this fact is ignored, and the generative algorithm is nonetheless successfully trained on discrete data. In the context of physics-based applications, this was done in \cite{liu2017simulating}, in which GANs were trained to simulate the 2d Ising model. However, in this work, we shall make use the probability density formulation of discrete spin-glasses introduced recently in \cite{hartnett2020probability} to convert the Boltzmann distribution over discrete spin variables into a probability density defined over continuous variables. A similar approach was used by \cite{li2018neural} to study the phase transition of the 2d Ising model using normalizing flows. 

In the probability density formulation of spin-glasses, discrete Boltzmann distributions of the form ${p(s) = e^{-\beta H}/Z_s}$, with $H$ given by Eq.~\ref{eq:Hspinglass}, are equivalently described in terms of a continuous random variable $x \in \mathbb{R}^N$, with probability density $p(x)$. The probability density may also be written as a Boltzmann distribution, i.e. $p(x) = e^{-\beta \mathcal{H}_{\beta}(x)}/Z_x$, where $\mathcal{H}_{\beta}(x)$ is the Hamiltonian \textit{density}, and the term density refers to how $\mathcal{H}_{\beta}(x)$ transforms under a change of variables \cite{hartnett2020probability}. The Hamiltonian density is given by
\begin{equation}
    \label{eq:H_of_x}
    \mathcal{H}_{\beta}(x) := \frac{1}{2} x^T \tilde{J} x - \frac{1}{\beta} \sum_{i=1}^N \ln 2\cosh\left( \beta \left( \tilde{J} x + h \right) \right)_i \,.
\end{equation}
We have added a subscript to the Hamiltonian density to emphasize its dependence on the inverse temperature. Here $Z_x = \int \mathrm{d}^N x \, e^{-\beta \mathcal{H}_{\beta}(x)}$ is the partition function of the $x$-variable, which is related to the original partition function of the $s$-variable via:
\begin{equation}
	\label{eq:partition_relation}
	Z_x  = (2\pi)^{N/2} (\det (\beta \tilde{J}))^{-1/2} e^{N \beta \Delta_/2} Z_s \,.
\end{equation}
The Hamiltonian density is written in terms of a shifted coupling matrix $\tilde{J}$, given by
\begin{equation}
	\tilde{J} := J + \Delta \, \mathbbold{1}_{N \times N} \,,
\end{equation}
where $\Delta := \max(0, \epsilon - \lambda_1(J) )$ with $0 < \epsilon \ll 1$. The shift $\Delta$ ensures that $p(x)$ is integrable. 

The variable $x$ may be interpreted as the integration variable in a somewhat non-standard application of the Hubbard-Stratonovich transformation. The probability density formulation of \cite{hartnett2020probability} (which builds off of the technique introduced in \cite{zhang2012continuous}) treats $x$ and $s$ as random variables with joint distribution $p(x, s)$ and conditional distributions $p(x|s)$, $p(s|x)$. The conditional distribution $p(x|s)$ is simply a multi-variate Gaussian:
\begin{equation}
    \label{eq:joint}
    p(x|s) = \mathcal{N}(\mu, \Sigma) \,,
\end{equation}
with mean $\mu = s$ and covariance matrix $\Sigma = (\beta \tilde{J})^{-1}$. The other conditional distribution is given by
\begin{equation}
    p(s|x) = \prod_{i=1}^N \frac{\exp\left(\beta s_i \left(\tilde{J} x + h\right)_i \right)}{2\cosh\left(\beta s_i \left(\tilde{J} x + h\right)_i \right)} \,.    
\end{equation}
Importantly, both conditional distributions are relatively easy to sample from. Thus, given a sample of the discrete spins $s$, a corresponding continuous configuration $x$ may be easily obtained, and vice versa.

\subsection{The Sherrington-Kirkpatrick model}
As mentioned in the introduction, the family of Boltzmann distributions with the Hamiltonian of Eq.~\ref{eq:Hspinglass} is very general. Therefore, as a first step towards answering the broader question of whether normalizing flows can successfully learn complex spin-glass distributions, we shall restrict our attention to one particular spin-glass, the SK model \cite{sherrington1975solvable}. The SK model is defined by specifying that the couplings $J_{ij}$ be drawn from an iid Gaussian distribution:
\begin{equation}
J_{ij} \sim \mathcal{N}\left( 0, \frac{\mathcal{J}^2}{N} \right) \,, \qquad (i < j) \,,
\end{equation}
where the $i > j$ values are fixed by the symmetry of $J$ to be the same as the $i < j$ values, and the diagonal entries are zero. In order to machine learn this complex distribution, we shall work in terms of the continuous variable $x$.

Despite its simple form, the SK model is a toy model with a very rich theoretical structure. In particular, there is a spin-glass phase transition at a critical temperature of $T_{\text{crit}} = \mathcal{J}$. Above this temperature the spin-glass exists in a disordered phase. For lower temperatures ergodicity is broken by the appearance of an exponentially large number of metastable states separated by large energy barriers. Thus, the SK model represents an excellent testing ground for modeling complex spin-glass distributions using deep neural networks.

A key result of \cite{hartnett2020probability} regarding the continuous formulation of spin-glass distributions is the existence of a convex/non-convex transition which occurs for the SK model at a temperature of
\begin{equation}
	T_{\text{convex}} := 4 \mathcal{J} \,.
\end{equation} 
Above $T_{\text{convex}}$ the Hamiltonian density is convex, and it becomes non-convex as the temperature is lowered past this value. Importantly, $T_{\text{convex}}$ is much higher than the spin-glass critical temperature. Thus, it is not obvious a priori whether the continuous formulation will be useful for the task of machine learning the SK model, since $p(x)$ becomes ``complex'' (i.e. has a non-convex energy landscape) well before the spin-glass phase appears.

\section{Normalizing flows and statistical physics \label{sec:normalizingflow_and_statmech}}
In this section, we introduce an application of real normalizing flows \cite{rezende2015variational} for modeling statistical many-body systems with real degrees of freedom. Normalizing flows are mappings from a latent random variable $z$ to a physical random variable $x$: $x = G(z)$. The latent variable is drawn from a simple prior distribution, for example a multivariate Gaussian. The flow $G$ transforms this into a new distribution $p_G(x)$, and in most machine learning applications the goal is then to modify the parameters of $G$ such that this induced distribution closely matches some target distribution. In our case, $x$ is the continuous spin variable introduced above, and the target distribution is the Boltzmann distribution $p(x) = e^{-\beta \mathcal{H}_{\beta}(x)}/Z_x$. We will argue below that the physical interpretation of $z$ is that it provides a coarse-grained description of the system. Denoting the prior as $p_z(z)$, the distribution induced by the flow can be obtained using the \textit{change of variables formula}
\begin{equation}
\label{eq:changeofvar_x}
p_{G}(x) := p_z(G^{-1}(x)) \left| \det \frac{\partial G^{-1}(x)}{\partial x} \right| \,,
\end{equation}
where the second factor on the RHS is just due to the Jacobian of the transformation. 

Real Non-Volume Preserving (NVP) flows \cite{dinh2016density} are normalizing flows for which the mapping $G$ is composed of multiple simpler invertible functions called affine layers, which we denote by $g^{(\ell)}$, and whose precise definition is given in Appendix \ref{sec:nvpflow}. Letting $L$ denote the number of such layers, the flow may be defined recursively as ${G_{(\ell)} := g_{(\ell)} \circ G_{(\ell-1)}}$, for $\ell = 1, ..., L$, with the understanding that $G_{(0)}$ is the identity and $G_{(L)}$ corresponds to the complete flow. The output of each layer of the flow represents a random variable $z_{(\ell)} := G_{(\ell)}(z_{(0)})$, with $z_{(0)} = z$ and $z_{(L)} = x$. 

The change of variables formula may be applied to each intermediate variable to obtain an expression for the associated probability density. For applications to statistical physics, it is useful to represent each of these as Boltzmann distributions. In other words, for each layer we introduce an associated Hamiltonian density $\mathcal{H}_{G_{(\ell)}}$ implicitly defined by
\begin{equation}
    p(z_{(\ell)}) := \frac{e^{-\mathcal{H}_{G_{(\ell)}}(z_{(\ell)})}}{Z_G} \,.
\end{equation} 
Taking the prior to be an isotropic Gaussian with zero-mean and standard deviation equal to one, the Hamiltonian density of the prior variable $z$ is just the quadratic function
\begin{equation}
    \mathcal{H}_{G_{(0)}}(z) = \frac{1}{2} z^T z \,.
\end{equation}
The change of variable formula may then be used to obtain $\mathcal{H}_{G_{(\ell)}}(z_{(\ell)})$ for any of the intermediate layers:
\begin{equation}
    \label{eq:internalH}
    \mathcal{H}_{G_{(\ell)}}(z_{(\ell)}) := \frac{1}{2} G_{(\ell)}^{-1} (z_{(\ell)})^T G_{(\ell)}^{-1}(z_{(\ell)}) - \ln \left| \det  \left( \frac{\partial G_{(\ell)}^{-1}(z_{(\ell)})}{\partial z_{(\ell)}}  \right) \right| \,.
\end{equation}
Similarly, the flow partition function may be expressed as an integral over any of the $z_{(\ell)}$:
\begin{equation}
    Z_G := \int \mathrm{d}^N z_{(\ell)} e^{-\mathcal{H}_{G_{(\ell)}}(z_{(\ell)})} \,.
\end{equation}
Evaluating this expression for $\ell = 0$, it is easily seen that $Z_G = (2\pi)^{N/2}$.

\begin{figure}[ht!]
  \centering
  \includegraphics[width=0.45\linewidth]{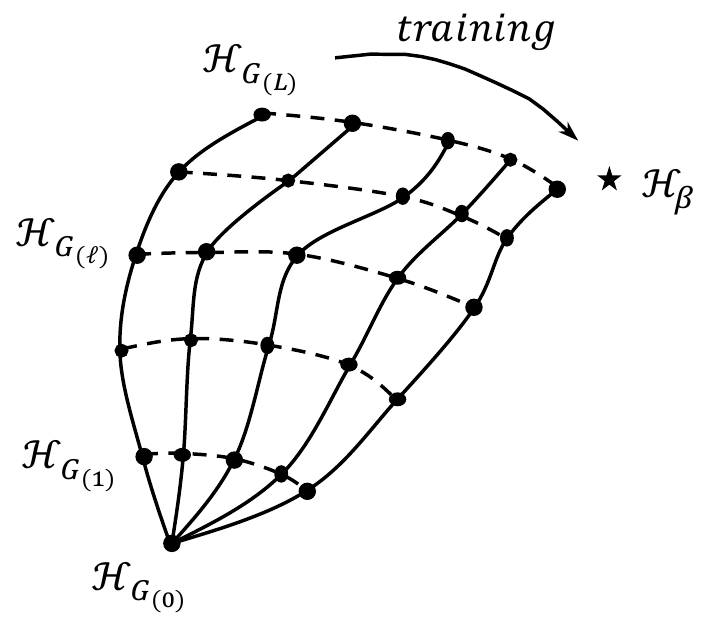}
 \captionof{figure}{\label{fig:flow_schematic} 
  Each layer of the flow can be interpreted as a physical system with an associated Hamiltonian density, $\mathcal{H}_{G_{(\ell)}}$. Forward evolution through the flow transforms $\mathcal{H}_{G_{(\ell)}}$ to $\mathcal{H}_{G_{(\ell+1)}}$, and thus the flow defines a discrete path through the space of Hamiltonian densities over $\mathbb{R}^N$. The effect of training is to modify this path so that the density of the final layer well approximates the true distribution $\mathcal{H}_{\beta}$.
}
\end{figure}

Describing the intermediate layers of the flow as random variables with their own Hamiltonian densities allows for a nice physical interpretation of the flow, see Fig.~\ref{fig:flow_schematic}. Initially, the sequence of flow Hamiltonian densities $\mathcal{H}_{G_{(\ell)}}(z_{(\ell)})$ are randomly initialized, and have no relation to the physical Hamiltonian density describing the system of interest $\mathcal{H}_{\beta}(x)$. The goal of training the normalizing flow is to modify the parameters of $G$ so that the final flow Hamiltonian $\mathcal{H}_{G_{(L)}}(z_{(L)})$ serves as a better and better approximation to the physical one. We will first investigate two distinct training methods based on different objective functions, and evaluate how well the trained models are able to capture the key properties of the spin-glass phase. Later, in Sec.~\ref{sec:internal}, we will present evidence that the trained flows implement a reversible coarse-graining on the physical system which is different than irreversible RG techniques. 

\section{Self-supervised learning of deep generative spin-glass models \label{sec:learning}}
Self-Supervised learning generally refers to a machine learning paradigm where all the labels in the training data can be generated autonomously \cite{Doersch_2017_ICCV}. In this paradigm, the architecture and the learning process could be fully supervised, however one does not need to manually label the training data. For example, the data can be labelled automatically by finding or computing the correlations between different input signals. Self-supervised learning is intrinsically suitable for online learning when there is steady stream of the new data that should be automatically digested on the fly, or whenever robots/agents need to automatically adapting within a changing environment, or whenever new instances of data or partial data can be generated by the future output of the model itself. Here, we use two different self-supervised learning approaches to train our deep generative spin-glass models using forward or backward KL divergence on the data that is automatically generated by either Monte Carlo sampling or by the model itself respectively.   

Generative models such as normalizing flows are typically trained by maximizing the log likelihood of some training data. Here, a priori we do not have any dataset, thus we need to generate our data; e.g.; through traditional Monte Carlo techniques such as parallel tempering \cite{PT_review}. Another major difference is that a tractable expression for the true probability distribution is known; i.e. the Boltzmann distribution over $x \in \mathbb{R}^N$ with Hamiltonian density $\mathcal{H}_{\beta}(x)$ is given by Eq.~\ref{eq:H_of_x}. This allows for two viable approaches towards training the flow to approximate the true distribution - one based on minimizing the reverse KL divergence and one based on minimizing the forward KL divergence.

\subsection{Reverse KL minimization \label{sec:reverse}}
In this approach the loss function is given by the reverse Kullback-Liebler (KL) divergence, plus a constant offset:
\begin{equation}
    \mathcal{L}_{\text{reverse}} := \mathbb{KL}\left( p_G(x) \left| \right| p(x) \right) - \ln Z_x  \,.
\end{equation}
The divergence is said to be ``reversed'' because in most machine learning applications the ``forward'' version $ \mathbb{KL}\left( p(x) \left| \right| p_G(x) \right)$
is used.\footnote{The KL divergence of two probability densities $p(x)$ and $q(x)$ is defined to be
$$ \mathbb{KL}(p(x) || q(x)) := \int \mathrm{d} x \, p(x) \ln\left( \frac{p(x)}{q(x)} \right) \,.$$
} This objective function was introduced as a way to train generative models in probability density distillation \cite{oord2017parallel}, and it was also used in \cite{li2018neural} to train normalizing flows to approximate the 2d Ising model near criticality. An important requirement for this approach is that both the true density $p(x)$ and the flow density $p_G(x)$ should be efficiently computable. The reverse loss function $\mathcal{L}_{\text{reverse}}$ has an important interpretation in statistical physics; it is proportional to the Gibbs free energy:
\begin{equation}
    G_x[p_G(x)] := \beta^{-1} \, \mathbb{KL}\left( p_G(x) \left| \right| p(x) \right) + F_x \,,
\end{equation}
where $F_x := -\beta^{-1} \ln Z_x$ is the Helmholtz free energy. The $x$ subscript is meant to indicate that these free energies are associated with the continuous spin-glass distribution $p(x)$, rather than the discrete one. Thus, minimization of the reverse KL loss is equivalent to the minimization of the Gibbs free energy. The global minimum is obtained when $p_G(x) = p(x)$, in which case the KL divergence vanishes and the Gibbs and Helmholtz free energies agree, $G_x[p(x)]=F_x$.

In order to employ this loss in a machine learning algorithm, we will make use of the fact that the reverse KL loss may also be written as an expectation over the prior:
\begin{align}
    \label{eq:reverseKL}
    \mathcal{L}_{\text{reverse}} 
    &= \mathbb{E}_{z \sim p_z(z)} \left( \beta \mathcal{H}_{\beta}(G(z)) - \mathcal{H}_G(G(z)) \right) \,.
\end{align}
Thus, rather than estimating a loss over a training set, with this training method the model itself generates a number of samples, and the loss is then estimated by computing the difference of the Hamiltonian densities (or equivalently, the log likelihoods). This therefore represents a potentially very powerful learning method, since the training is limited only by computational constraints, and not by the quantity or quality of data. Since the KL divergence is non-negative, this loss provides a lower bound for the loss function/Gibbs free energy:
\begin{equation}
\label{eq:reverse_loss_bound}
\mathcal{L}_{\text{reverse}} = \beta \, G_x[p_G(x)] \ge \beta F_x \,.
\end{equation}
The tightness of this bound provides one measure of how well the flow has learned the spin-glass distribution. 

\subsection{Forward KL minimization \label{sec:forward}}
A more conventional approach towards training the normalizing flow is to first assemble an unlabeled dataset $\mathcal{D} = \{ x^{(i)} \}_{i=1,...,|\mathcal{D}|}$ consisting of spin-glass configurations distributed according to the Boltzmann distribution. We will generate such a dataset by performing Monte Carlo sampling of the true spin-glass distribution $p(x)$. The loss function will then be taken to be the expectation of the negative log-likelihood over the dataset:
\begin{equation}
    \label{eq:forwardKL_1}
    \mathcal{L}_{\text{forward}} := - \mathbb{E}_{x \sim p(x)} \ln p_G(x) \,.
\end{equation}
which is equivalent to the forward KL divergence plus the Shannon entropy of $p(x)$:
\begin{equation}
    \label{eq:forwardKL_2}
    \mathcal{L}_{\text{forward}} = \mathbb{KL}\left( p(x) || p_G(x) \right) + H(p(x)) \,.
\end{equation}
$\mathcal{L}_{\text{forward}}$ may also be written as the the cross entropy between $p(x)$ and $p_G(x)$. Similar to the reverse KL loss function, the KL loss function is lower-bounded by the Shannon entropy,
\begin{equation}
    \label{eq:forward_loss_bound}
    \mathcal{L}_{\text{forward}} \ge H(p(x)) \,.
\end{equation}

\subsection{Training deep generative spin-glasses with normalizing flows}
We considered the problem of training normalizing flows to approximate the SK model for $N = 256$ spins\footnote{This is a small number of spins by some standards - for example it represents a $16 \times 16$ square lattice in two dimensions. However, recall that the SK model is defined over a complete graph, which in this case contains 32,640 bonds between spins. For context, a $16\times 16$ square lattice with periodic boundary conditions and $N = 16^2 = 256$ spins contains only 512 bonds.} and $0 < T \le 5 \mathcal{J}$, with the range of temperatures chosen to encompass both the spin-glass and paramagnetic phases of the SK model. This range extends far beyond the phase transition temperature of $T_{\text{crit}} = \mathcal{J}$ so that we may study the effect of the transition between a convex and non-convex Hamiltonian density, which occurs for $T_{\text{convex}} = 4 \mathcal{J}$. For each temperature, we considered 10 different realizations of the disorder (i.e. 10 random draws of the matrix $J_{ij}$). Also, we set $\epsilon = 0.01$, so that the smallest eigenvalue of the shifted coupling matrix is $\lambda_1( J_{\Delta}) = 0.01$ (which effectively adjusts $T_{\text{convex}}$ to be $4.01 \mathcal{J}$).

We performed experiments for both training methods discussed above, minimizing either the reverse KL divergence or the forward KL divergence. For the forward KL approach, we first assembled a dataset $\mathcal{D}$ of spin configurations using parallel-tempering (PT) (see for example \cite{katzgraber2010introduction}). We  simulated 20 parallel systems (called replicas), each at a different temperature. We recorded the spin-configurations throughout the parallel tempering after each sweep of all $N$ spins, until 100,000 samples had been obtained for each temperature. We then repeated this procedure for 10 realizations of the coupling matrix $J_{ij}$. Lastly, to train the normalizing flow the discrete Ising spin configurations must be converted to continuous configurations. We used each spin configuration $s$ to generate a single $x$ configuration using the joint distribution $x \sim p(x|s)$ (given by Eq.~\ref{eq:joint}).\footnote{Since the variables $s$ and $x$ are related probabilistically, each $s$ configuration could have been used to generate any number of $x$ configurations. We chose a 1:1 conversion ratio for simplicity.}

\begin{figure*}[ht!]
  \centering
  \includegraphics[width=0.8\linewidth]{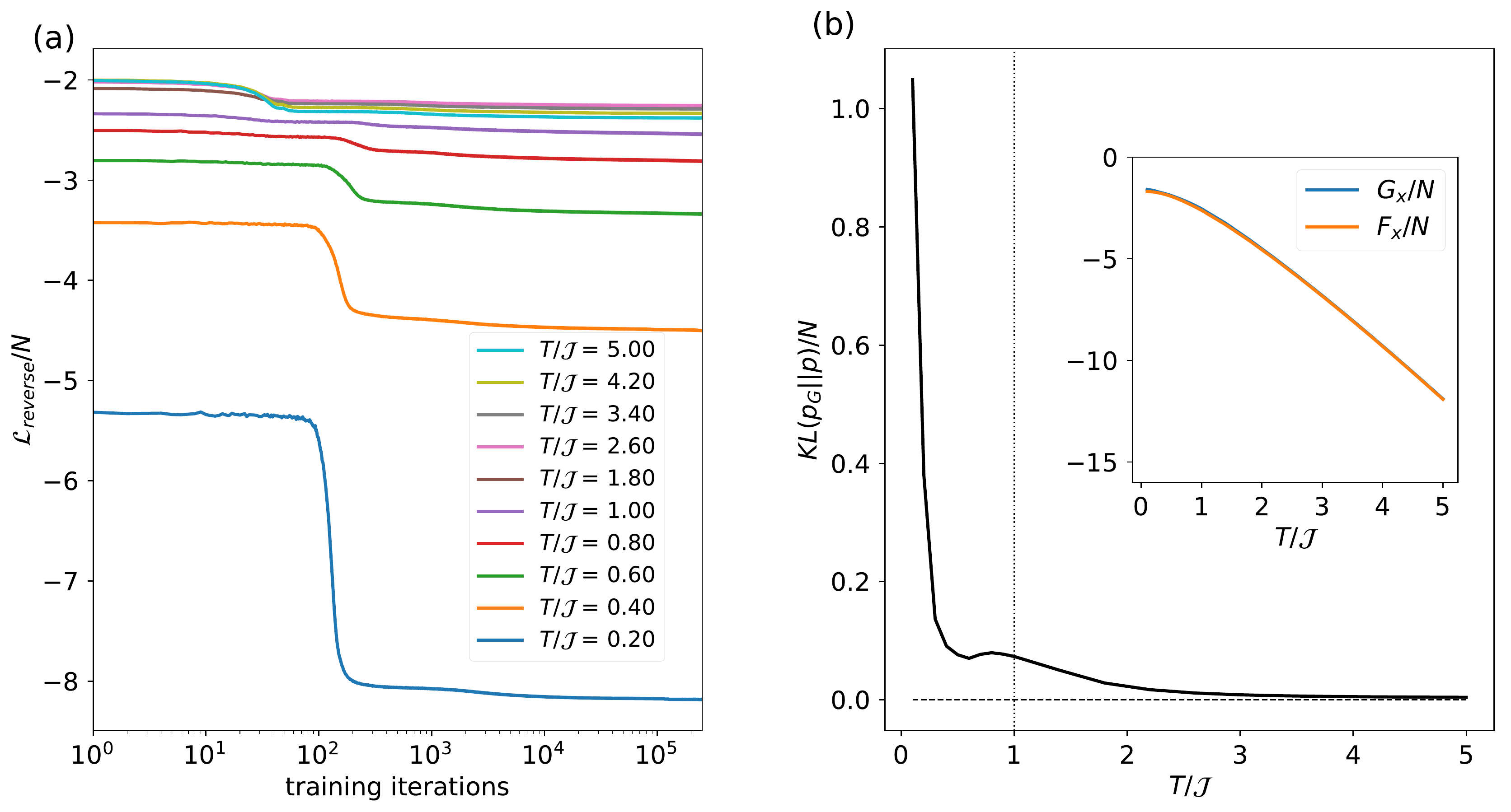}
 \captionof{figure}{\label{fig:nvp_loss_reverse} 
  (a) The reverse KL loss $\mathcal{L}_{\text{reverse}}$ measured throughout the training. Each curve represents an average over 10 experiments, each with a different draw of the coupling matrix $J_{ij}$.
  (b) The estimated value of the reverse KL divergence as a function of $T/\mathcal{J}$. The inset shows both the Gibbs and Helmoltz free energies. The closeness of the Gibbs and Helmholtz free energies in (b) is an indication that the flow has been successfully trained. However, estimating the KL divergence $\mathbb{KL}(p_G(x) || p(x)) = \beta(G_x - F_x)$ shows that the quality of the normalizing flow approximation to the true Boltzmann distribution nonetheless becomes quite poor at low-temperatures. This is an indication that reverse KL-trained flow is not capturing key properties of the spin-glass phase at low-temperatures.
}
\end{figure*}

In Fig.~\ref{fig:nvp_loss_reverse} (a) we plot the reverse KL loss throughout the course of training. The loss is roughly constant for the first 100 iterations, after which it drops sharply. For $T/\mathcal{J} < T_{\text{crit}}$ it then continues to decrease slowly with additional training. The magnitude of the drop increases for low temperatures. A measure of the quality of the trained flow may be obtained by examining the tightness of the free energy bound in Eq.~\ref{eq:reverse_loss_bound}, or equivalently, the smallness of the reverse KL divergence. In order to estimate the KL divergence, we first evaluated the Helmholtz free energy $F_x$ using the relation between $Z_x$ and $Z_s$ in Eq.~\ref{eq:partition_relation}. In the spin-glass phase $(T < T_{\text{crit}})$ the discrete partition function $Z_s$ was estimated using the results of the parallel tempering described above, while in the paramagnetic phase ($T > T_{\text{crit}}$) we used the replica-symmetric expression $F_s = - N \beta \mathcal{J}^2/4 - N \ln 2/\beta$, where $F_s := - \beta^{-1} \ln Z_s$. Importantly, the replica-symmetric expression is exact in the paramagnetic phase and in the large-$N$ limit. Fig.~\ref{fig:nvp_loss_reverse} (b) shows that the reverse KL divergence is rather high at low temperatures, and decreases rapidly as the temperature is raised. The inset shows the estimated value of the two free energies, Gibbs and Helmholtz. 

\begin{figure*}[ht!]
  \centering
  \includegraphics[width=0.8\linewidth]{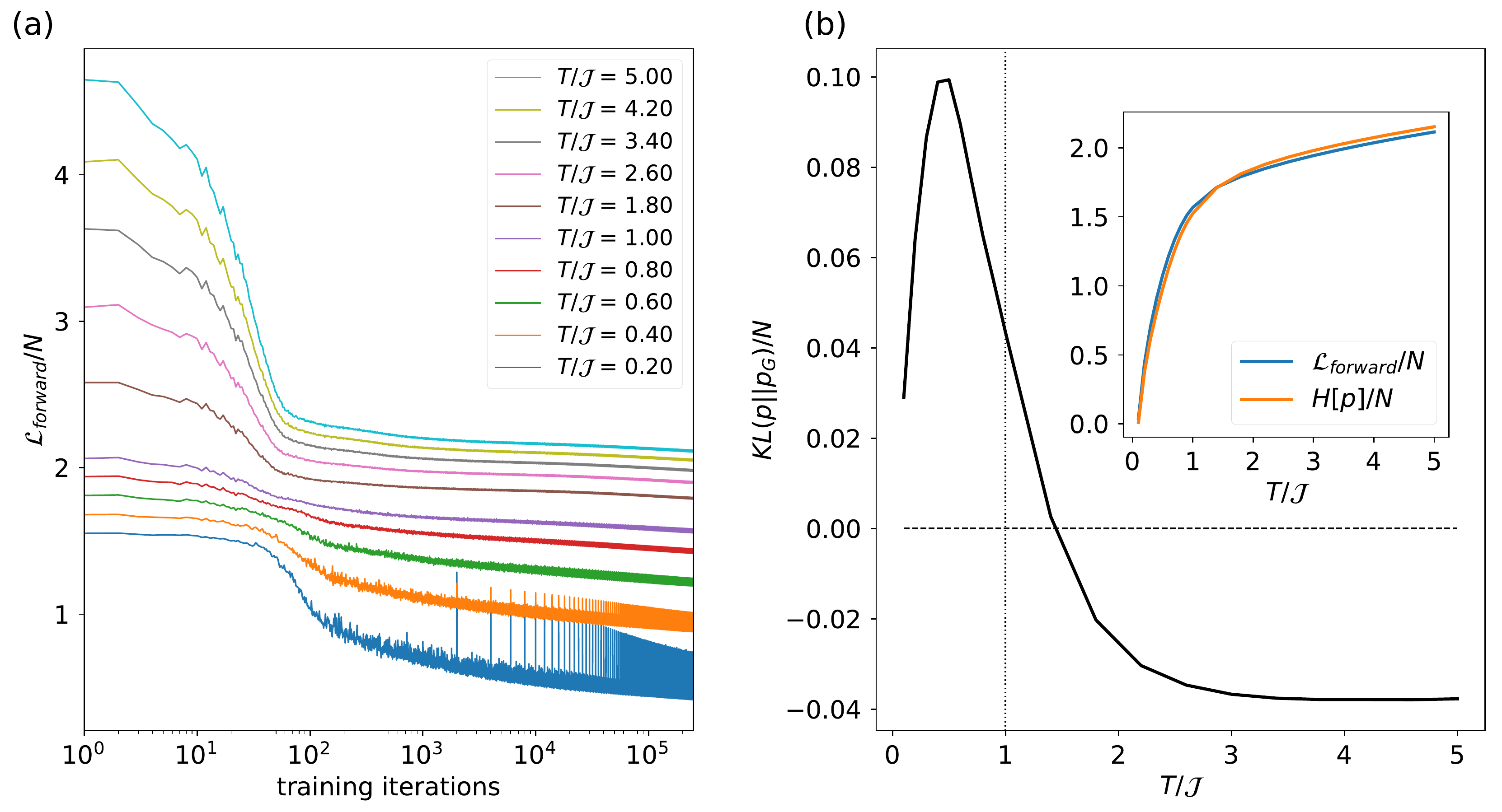}
 \captionof{figure}{\label{fig:nvp_loss_forward} 
 (a) The forward KL loss $\mathcal{L}_{\text{forward}}$ measured throughout the training. Each curve represents an average over 10 experiments. 
 (b) The estimate of the forward KL divergence as a function of $T/\mathcal{J}$. The inset depicts the forward loss $\mathcal{L}_{\text{forward}}$ and the (normalized) Shannon entropy $H[p(x)]/N$. The closeness of the forward loss and the Shannon entropy in (b) is an indication that the flow has been successfully trained. Correspondingly, the estimated KL divergence is quite small for the entire range of temperatures studied, although it does experience a peak at $T=0.5$. 
}
\end{figure*}

Similarly, in Fig.~\ref{fig:nvp_loss_forward} (a) we plot the forward KL loss throughout the course of training.  In the spin-glass phase, learning proceeds slowly at first 100 iterations or so, and then decreases much more rapidly afterwards. In the paramagnetic phase, the loss decreases quickly with the first 100 iterations, and then decreases much more slowly afterwards. For the critical temperature, $T = T_{\text{crit}}$, the loss curve is essentially flat, and there is not a period of steep decline. Another interesting observation is that the variation in the value of the loss increases as $T$ decreases. The data for the lowest temperature considered, $T/\mathcal{J} = 0.2$, shows that after about 1000 training iterations there are occasionally large fluctuations in the value of the loss.

As with the reverse KL loss function, the quality of the learned flow can be measured by estimating the KL divergence and the tightness of the bound Eq.~\ref{eq:forward_loss_bound}. This is depicted in Fig.~\ref{fig:nvp_loss_forward} (b), and the inset compares the asymptotic value of the forward KL loss with the Shannon entropy of the physical distribution $p(x)$, which we evaluated using the results of the parallel tempering simulations. Compared to the reverse KL divergence, the forward KL divergence first increases rapidly as the temperature is raised, reaches a peak at about $T =0.5$, then decreases rapidly across the spin-glass phase transition, and finally approaches an asymptotic value as the temperature is increased further. The forward KL divergence achieves much smaller values than the reverse KL divergence in the spin-glass phase - the maximum value of the forward KL divergence is an order of magnitude smaller than the maximum value of the reverse KL divergence. One important point is that the estimate of the forward KL divergence actually becomes negative for high temperatures. As the KL divergence is non-negative, this indicates that the error in the estimate is large enough to change the value from positive to negative. We initially observed a similar phenomenon in estimating the reverse KL loss, but we were able to improve our estimate in the paramagnetic phase by using the exact replica symmetric expression for the free energy. We are not aware of an analogous expression for the Shannon entropy $H[p(x)]$.

\subsection{Order parameter of generative spin-glasses}
Spin-glasses are characterized by the presence of a large number of metastable states. These are typically separated from one another by large energy barriers, and as a result ergodicity is broken at low temperatures and the system will become stuck in one of the many states. The closeness of two such states $\alpha$ and $\beta$ may be measured by the overlap between them:
\begin{equation}
    q_{\alpha \beta} := \frac{1}{N} \sum_{i=1}^N m_i^{(\alpha)} m_i^{(\beta)} \,.
\end{equation}
Here, $m_i^{(\alpha)}$ is the average per-site magnetization, or marginal probabilities, within the state $\alpha$, and similarly for $m_i^{(\beta)}$. As shown by Parisi, the distribution of overlap values $P_J(q)$ for a given spin-glass instance may be used as an order parameter for the spin-glass phase \cite{parisi1980order}. This represents a key distinction from more conventional phases of matter where the order parameter is simply a real number - here, it is an entire \textit{function}. Equivalently, the order parameter for a spin-glass could be considered to be the infinite collection of real numbers defined by the moments of the overlap distribution. This unique nature of the spin-glass order parameter underscores the complexity of these systems compared to more conventional phases of matter.

\begin{figure*}
  \centering
  \includegraphics[width=0.98\linewidth]{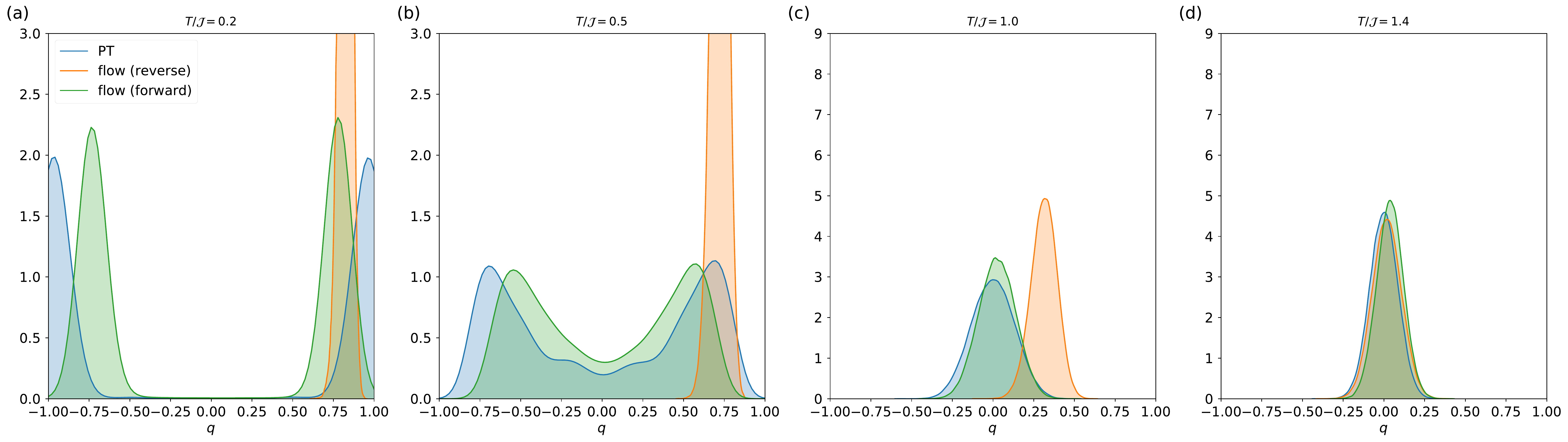}
 \captionof{figure}{\label{fig:overlap} 
 The overlap distribution for samples of $p(s)$ generated using parallel tempering (blue), samples generated from the reverse KL-trained flow (orange), and samples generated from the forward KL-trained flow (green). All results correspond to a single disorder realization. As the temperature is raised above $T_{\text{crit}} = \mathcal{J}$, the true overlap distribution should experience a transition from being bimodal to unimodal. Both the PT and forward KL-trained flow exhibit this transition, whereas the reverse KL-trained flow fails to capture this phenomenon as it has suffered from mode collapse.
}
\end{figure*}

In order to evaluate the extent to which the trained normalizing flows have successfully approximated the spin-glass phase we examined the overlap distribution. $P_J(q)$ may be approximated by sampling the spin-glass and computing the configurational overlaps between two randomly chosen spin configurations $s^{(a)}$ and $s^{(b)}$:
\begin{equation}
    q_{ab} := \frac{1}{N} \sum_{i=1}^N s_i^{(a)} s_i^{(b)} \,.
\end{equation}
As sample size grows, the distribution of configurational overlaps $q_{ab}$ will approach the overlap distribution $P_J(q)$. In Fig.~\ref{fig:overlap} we plot numerical estimates of the non-disorder-averaged overlap distribution $P_J(q)$ for the parallel tempering data and samples obtained from both the forward and reverse KL-trained flows.\footnote{The flow generates samples of the continuous variable $x$; samples of the discrete spin configuration $s$ were then obtained by sampling from the conditional distribution $s \sim p(s|x)$.} Neither the reverse or forward-trained flow perfectly matches the PT result, although the forward and PT results qualitatively match. The fact that the forward-trained flow is able to produce a bi-modal overlap distribution demonstrates that it has captured an important characteristic of the spin-glass phase. 
In contrast, the overlap distribution for the reverse KL-trained flow has just a single peak near $q=1$ for $T< T_{\text{crit}}$, which indicates that most of the samples are very close to one another. This suggests that the flow has not successfully modeled the key property of the spin-glass phase, namely the existence of many metastable states. Further evidence for this conclusion comes from examining the magnetization, $M = \sum_i \langle s_i \rangle$. We find that $M \approx 0$ for all sampling methods, but the reverse KL-trained flow is unique in that it produces non-zero per-site magnetizations, i.e. $\langle s_i \rangle \neq 0$. This is consistent with the hypothesis that the flow distribution has collapsed onto just a single mode of the true distribution. Indeed, this is a known failure mode of models trained on the reverse KL divergence.\footnote{
A nice discussion on this point may be found in Sec. 21.2.2 of \cite{murphy2012machine}.} 

\subsection{Ultrametricity in generative spin-glasses}
One of the most remarkable properties of spin-glasses is that their state space can be endowed with highly non-trivial topological structure, which emerges naturally from seemingly unstructured Hamiltonians. For instance, the simple SK model Hamiltonian, with a featureless fully connected underlying graph and iid random couplings, can be shown to generate a simple hierarchical structure in its solution space. Specifically, the low energy states of the SK model obey an interesting relation known as ultrametricity \cite{mezard1984nature}. Given any triplet of states, $a$, $b$, and $c$, the 3 possible overlaps $q_{ab}$, $q_{ac}$ and $q_{bc}$ may be converted into a set of Hamming distances $d_{ab} := (1-q_{ab})/2$ which form the 3 sides of the triangle connecting the states. The standard triangle inequality is $d_{ac} \le d_{ab} + d_{bc}$, and in the SK model, these distances are known to obey the stronger inequality $d_{ac} \le \max(d_{ab}, d_{bc})$ which defines ultrametric spaces. This condition implies some geometric properties, for example that all triangles are either equilateral or acute isosceles. The distribution of triangles may be computed analytically in the SK model using the the replica approach with Parisi's replica symmetry breaking ansatz. One finds that in the spin-glass phase 1/4 of the triangles are the equilateral, and 3/4 are acute isosceles.\footnote{By isosceles, here we mean those triangles which are isosceles and not also equilateral, since all equilateral triangles are also trivially isosceles.} 

\begin{figure*}
  \centering
  \includegraphics[width=0.98\linewidth]{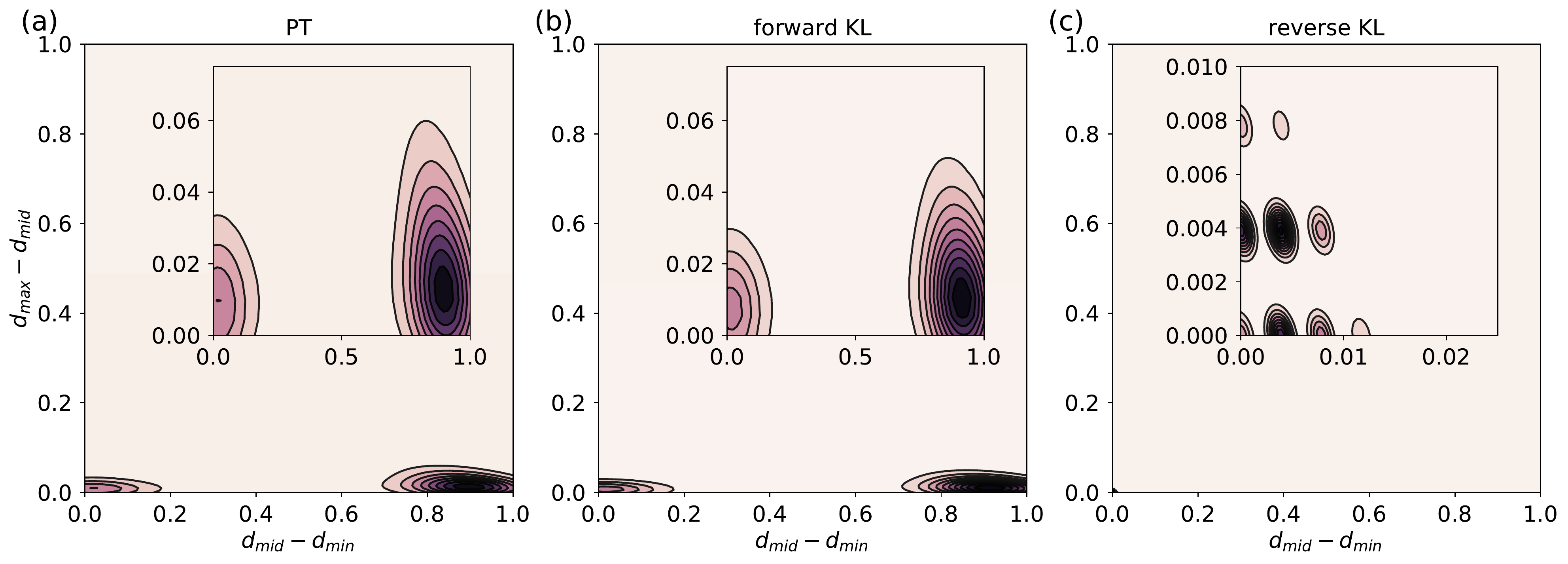}
 \captionof{figure}{\label{fig:ultrametric} The distribution of distances for $T/\mathcal{J} = 0.2$ for samples generated from a single disorder realization using 
 (a) parallel tempering, 
 (b) the forward KL-trained flow, and
 (c) the reverse KL-trained flow.
 The cluster at the origin corresponds to equilateral triangles, and the other cluster on the x-axis corresponds to acute isosceles triangles.
All triangles in ultrametric spaces are either equilateral or acute isosceles, and so the two well-separated clusters in (a) and (b) indicate that both the PT simulations and the forward KL-trained flows have correctly reproduced the ultrametric property known to be obeyed by the SK model. The requirement of ultrametricity severely restricts the topology of the state space, and this is therefore a non-trivial test of the learned normalizing flow. In contrast, (c) shows that that reverse KL-trained flow has completely failed to capture this property.
}
\end{figure*}

Therefore, as a final test of the quality of our learned continuous spin-glasses trained with normalizing flows, it would be essential to examine if such models can actually generate the emergent hierarchical/ultrametric structure of the SK Hamiltonian. We plot the distribution of triangle side lengths in Fig.~\ref{fig:ultrametric}. Letting $d_{\text{min}}, d_{\text{mid}}, d_{\text{max}}$ denote the ordered distances, the cluster near the origin of Fig.~\ref{fig:ultrametric} corresponds to equilateral triangles and the other cluster on the $x$-axis corresponds to acute isosceles triangles. This same plot was used in \cite{hartnett2018replica} in the context of the bipartite SK model. Fig~\ref{fig:ultrametric} (a) depicts the triangle distribution for samples generated by parallel tempering simulations, and Fig.~\ref{fig:ultrametric} (b) depicts the forward KL-trained flow distribution. These are in close agreement with one-another, and both exhibit two well-separated clusters of equilateral and isosceles triangles. Moreover, in both cases the data is very close to the 1:3 theoretical ratio of equilateral to isosceles triangles. In contrast, the samples generated from the reverse KL-trained flow depicted in Fig.~\ref{fig:ultrametric} (c) form several isolated clusters very near the origin, which is a manifestation of the mode collapse in this generative model. 

\section{A phase transition within the internal layers of the normalizing flow \label{sec:internal}}

\begin{figure*}
  \centering
  \includegraphics[width=0.98\linewidth]{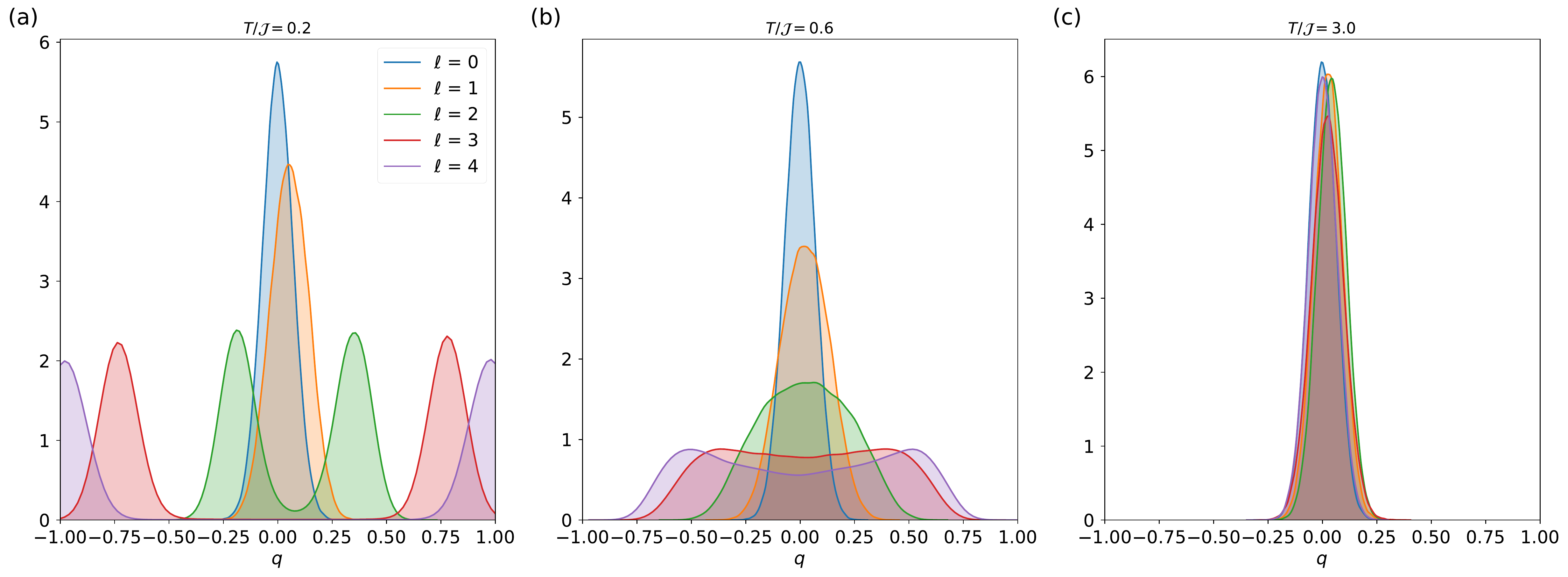}
 \captionof{figure}{\label{fig:overlaps_internal} 
 The (non-disorder averaged) overlap distribution $P_J(q)$ computed from sampling the internal layers of the flow. The $\ell = 0$ layer corresponds to just sampling the Gaussian prior, and the $\ell = L$ layer corresponds to standard sampling of the final layer of the flow ($L=4$ was used in our experiments). The flow was trained on parallel tempering samples of the SK model at various temperatures, corresponding to plots (a)-(c). In the low-temperature, spin-glass phase, the first layers of the flow produce a unimodal overlap distribution, and evolution through the flow transforms the final layers into a bi-modal distribution characteristic of the spin-glass phase. This is the evidence of a spin-glass transition within the layers of the flow itself. In contrast, in temperatures above the spin-glass phase all layers produce a unimodal distribution and there is no spin-glass order.
}
\end{figure*}

\begin{figure*}
  \centering
  \includegraphics[width=0.98\linewidth]{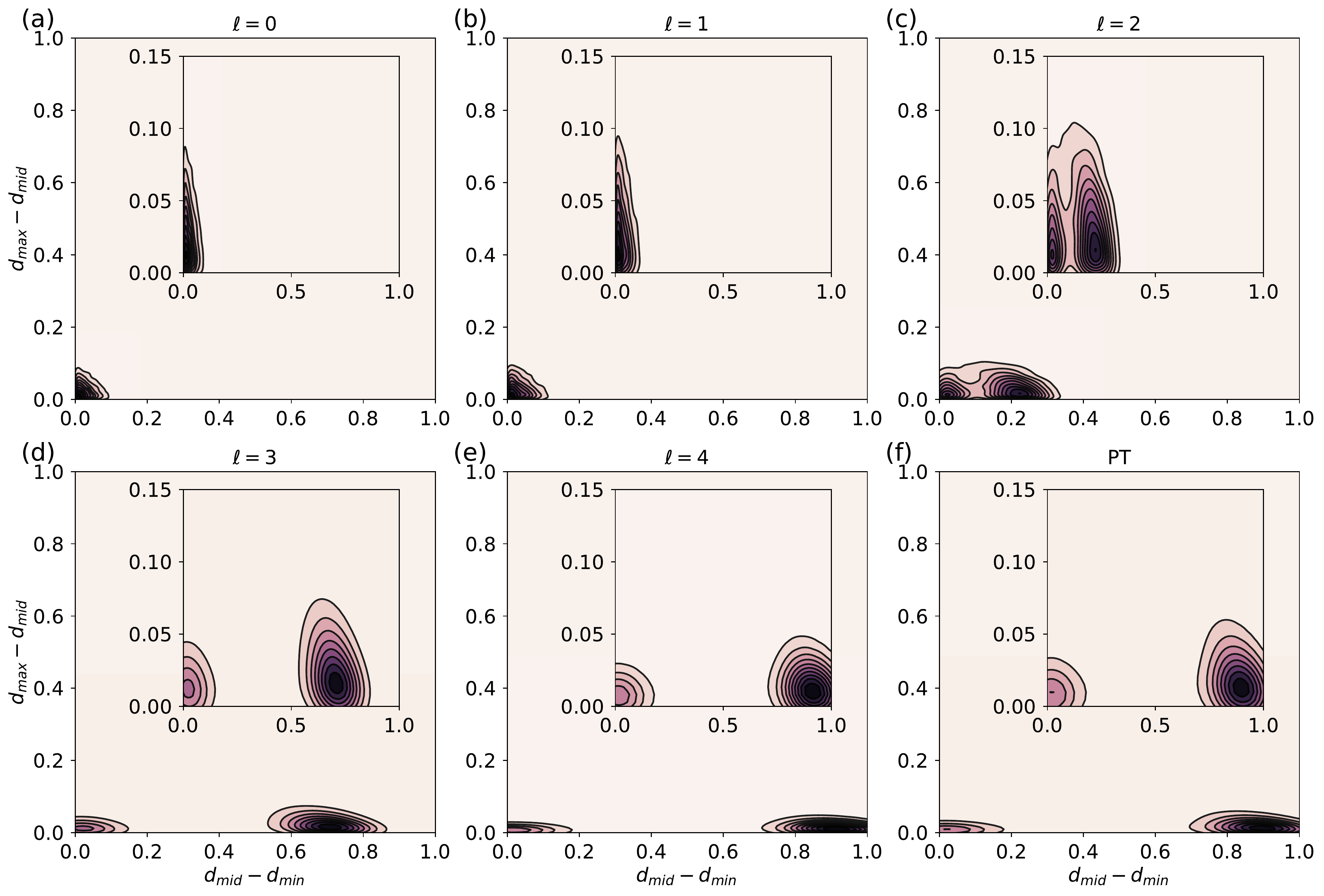}
 \captionof{figure}{\label{fig:ultrametric_internal} 
 (a) - (e) The triangle distance distribution for low energy states for each internal layer of a normalizing flow trained using the forward KL loss function, for $T/\mathcal{J} = 0.2$.
 (f) The same distribution, plotted for samples generated using parallel tempering applied to the original discrete spin distribution $p(s)$. All the plots correspond to a single disorder realization. The triangle distribution of the final layer (e) matches the distribution obtained from PT sampling (f) (as was shown in Fig.~\ref{fig:ultrametric}). (a) depicts triangle distribution produced by the Gaussian prior, and (b-d) show how forward evolution through the flow transforms this to a triangle distribution that closely resembles that one obtained by PT.
}
\end{figure*}

We have demonstrated that normalizing flows may be successfully trained to learn the spin-glass phase of the SK model. Recall that the flow induces a sequence of Hamiltonian densities $\mathcal{H}_{G_{(\ell)}}(z_{(\ell)})$ which interpolate from a Gaussian energy landscape at $\ell = 0$ to the complex, multi-modal landscape that corresponds to the spin-glass phase for $\ell = L$. This suggests that there is a phase transition \textit{within} the flow, e.g., in passing from layer $\ell$ to layer $\ell + 1$. 

In order to confirm this conjecture, $\ell$ would ideally be a continuous parameter which would allow for a critical layer $\ell_c$ to be found, and then critical exponents could be calculated with respect to $(\ell - \ell_c)/L$. One way to achieve this scenario would be to take $L \rightarrow \infty$, so that the normalizing flow is composed of an infinite number of layers, each of which implements an infinitesimal diffeormorphism. This would allow for additional theoretical analysis but would diminish the practical relevance since it is unclear how the abstracted normalizing flow with infinite layers relates to the finite-$L$ trained models considered above. An interesting alternative would be to use a form of flow based on a 1-parameter diffeomorphism such as Neural Ordinary Differential Equations (Neural ODEs) \cite{chen2018neural}. In lieu of taking the $L \rightarrow \infty$ limit or considering a different class of generative models, we will simply investigate the statistics of the internal layers of our trained flows and establish that, as $\ell$ increases, there appears to be a transition from the paramagnetic phase to the spin-glass phase. 

The Boltzmann distributions corresponding to the intermediate layers of the flow may be sampled by first sampling the Gaussian prior, and then only partially evolving the flow up to the corresponding intermediate layer. Following this approach, in Fig.~\ref{fig:overlaps_internal} we plot the sequence of overlap distributions $P_J(q)$ for each internal layer of the flow for 3 distinct temperatures. Fig.~\ref{fig:overlaps_internal} (a) shows that deep in the spin-glass phase the overlap distribution is transformed from the uni-modal distribution of the Gaussian prior to a bi-modal distribution in going from $\ell = 1$ to $\ell = 2$. Each additional layer then increases the separation of the two modes as the flow distribution becomes a closer approximation to the physical distribution. Fig.~\ref{fig:overlaps_internal} (b) shows that increasing the temperature (but keeping within the spin-glass phase) causes the uni-modal to bi-modal transition to occur deeper in the flow. Lastly, Fig.~\ref{fig:overlaps_internal} (c) demonstrates that in the paramagnetic phase the overlap distribution of \textit{all} layers is uni-modal, and there is no sign of a phase transition within the flow.

To gain more insight into the nature of the transition that occurs as $\ell$ is increased within the flow, in Fig.~\ref{fig:ultrametric_internal} we plot the triangle distance distribution for each internal layer for $T/\mathcal{J} = 0.2$. As in Fig.~\ref{fig:overlaps_internal}, in going from $\ell = 1$ to $\ell = 2$ the distribution appears to qualitatively change, and the triangle distribution becomes bi-modal. In this case the two modes correspond to equilateral and acute isosceles triangles. In the paramagnetic phase (and in the strict large-$N$ limit), there are only equilateral triangles, whereas the spin-glass phase is characterized by both equilateral and acute isosceles. Taken together, the results of Fig.~\ref{fig:overlaps_internal} and Fig.~\ref{fig:ultrametric_internal} show that the flow transforms a quadratic energy landscape into a spin-glass landscape via an incremental process, with each affine layer bringing the normalizing flow distribution closer to the true spin-glass distribution.

\section{Discussion and outlook \label{sec:discussion}}
We have demonstrated that real NVP flows can be successfully trained to model complex spin-glass distributions. We evaluated two methods for training the flow, one based on minimizing the Gibbs free energy, or reverse KL divergence, and another based on minimizing the forward KL divergence. Although the reverse KL approach led to a relatively tight bound for the free energy, which suggested that the model was successfully trained, the reverse KL divergence is quite large at low temperatures, and moreover the model was found to suffer from serious mode collapse. In contrast, the forward KL-trained model was able to capture the key properties of the spin-glass low-energy distributions, including a bi-modal overlap distribution and topological structure in the state space known as ultrametricity.

Normalizing flows are composed of a series of continuous changes of variables (i.e. diffeomorphisms), and thus they seem ill-suited for studying discrete systems such as spin-glasses directly. For this reason, as a first step before applying the normalizing flows, we used the probability density formulation recently introduced by us in Ref. \cite{hartnett2020probability} to convert the discrete Boltzmann distribution into a physically equivalent continuous one. This dual continuous distribution exhibits an important transition at $T_{\text{convex}} = 4 \mathcal{J}$ for the SK model. As this temperature is crossed from above, the Hamiltonian density $\mathcal{H}_{\beta}(x)$ transitions from being convex to being non-convex. As part of this transition, the critical point at $x=0$ becomes unstable and a pair of new critical points with $x \neq 0$ appears. Because this transition occurs at temperatures well above the spin-glass transition, one might suspect that the loss of convexity might hinder the learning task for approximating the spin-glass phase. However, as we have showed in \cite{hartnett2020probability} the spin-glass transition remains intact under continuous transformation and here we find no signature of any nonphysical effects when we use standard loss function based on forward KL divergence. Thus the continuous formulation can be used for spin-glass density estimation and generative modeling.\footnote{Other recent approaches for applying normalizing flows to discrete data may be found in \cite{hoogeboom2019integer, tran2019discrete}. It would be interesting to investigate how these approaches compare to the continuous relaxation approach of \cite{zhang2012continuous}.}

Compared to traditional machine learning settings, the reverse KL minimization approach seems appealing because it involves minimizing a natural cost function, i.e., the Gibbs free energy, and the amount of training data is essentially unlimited. Unfortunately, as we observed here, the quality of the self-generated ``data'' was not good enough to yield a reliable performance. In the forward KL approach the amount of data was finite (100,000 samples), although this is an artificial bound because we generated the data ourselves through traditional Monte Carlo approaches. This observation suggests an online extension of the forward KL learning where the Monte Carlo data generation occurs in tandem with the gradient-based optimization. Rather than generating a fixed dataset of Monte Carlo samples, a better approach would be to continuously sample from the parallel tempering replicas, and to periodically perform a gradient descent update for the normalizing flow once enough samples have been collected to form a new mini-batch. This would eliminate the potential for over-fitting since for each gradient descent update the mini-batch would consist of previously-unseen Monte Carlo samples.

Real NVP flows implement a sequence of changes of variables, also known as diffeomorphisms. An important mathematical result along these lines is that any two distributions over $\mathbb{R}^N$ can be related by a diffeomorphism. However, it is a non-trivial problem to find such a transformation, especially for spin-glasses, which are notoriously complex. Our successfully trained flow implements a very particular sequence of diffeomorphisms - the inverse flow $G^{-1}$ maps the the energy landscape of a spin-glass to the simple quadratic function $\frac{1}{2} z^T z$. Because a rough and multi-modal landscape is mapped to a quadratic one, we can think of $G^{-1}$ as a coarse-graining. Importantly, the coarse-graining is with respect to the ruggedness of the energy landscape - no degrees of freedom have been integrated out, and no information has been lost because the flow is invertible. This resembles simulated annealing \cite{Kirkpatrick671} and adiabatic quantum computation \cite{Lidar18} for discrete optimization in which one starts with significant thermal or quantum fluctuations, smoothing the energy landscape to be originally convex, and then gradually (adiabatically) transform it to a non-trivial noncovex distribution via very slow (exponential in the worse-case instances) passage through a thermal or quantum phase transition respectively. In the case of adiabatic quantum annealing the process is in principle deterministic and fully reversible. However, in practice one has to deal with finite-temperature dissipations, many-body quantum localization, and Griffiths singularities for strongly disordered quantum spin glasses in low dimensions \cite{Boixo16,Mohseni18}.   

There is also an apparent strong connection between our work and the renormalization group techniques. One might attempt to formally identify the inverse normalizing flow with a renormalization group flow. However, we are not aware of any formulation of the renormalization group which applies directly to probability densities such as the ones considered here. A possible connection of between RG and normalizing flows can be done through the Neural ODEs representations \cite{chen2018neural}. The renormalization of spin-glasses appears to be most well-studied for hierarchical models, and thus these would be a natural starting point to develop this connection further.\footnote{The thesis \cite{castellana2013renormalization} contains a good overview of the renormalization of these systems.} 

Some years ago an exact mapping between a variational RG transformation and Boltzmann Machines (BMs) with hidden layers was proposed in \cite{mehta2014exact}. There are clearly some similarities between their setup and ours. In our work, the internal layers of the generative model are drawn from Boltzmann distributions with Hamiltonian densities $\mathcal{H}_{G_{(\ell)}}(z_{(\ell)})$, and moreover these internal distributions may be easily sampled. In contrast, in Ref. \cite{mehta2014exact} BMs were studied over discrete variables, whose internal layers are hard to sample from. Perhaps the most significant difference between their work and ours is that in our case the marginal distribution of the zeroth layer is constrained to be a simple Gaussian, whereas for BMs the marginal distribution over the final hidden layer is intractable and depends on the weights in a very non-trivial way. The fact that in normalizing flows the distribution over the latent variable is fixed to be a Gaussian means that the inverse flow may be interpreted as a coarse-graining map, since the endpoint of the inverse flow is always a quadratic energy landscape.

Further developing the connection between the renormalization group and normalizing flows would be worthwhile since, unlike traditional coarse-graining or renormalization group approaches which often rely on symmetries or other assumptions not likely to be found in systems of practical interest, the coarse-graining $G^{-1}$ is separately \textit{learned} for each spin-glass system, making our approach broadly applicable. In particular, it has been shown in Ref. \cite{li2018neural} that the coarse-graining performed by the inverse flow may be used to improve Hamiltonian Markov Chain sampling of the system. Normalizing flows have also been recently used to accelerate sampling of a simple lattice quantum field theory \cite{albergo2019flow}. The most important practical application of our work would be to obtain similar improvements in the computational tasks associated with spin-glasses, such as combinatorial optimization, sampling of multi-modal distributions, and inference in graphical models.

\subsubsection*{Acknowledgments}
We would like to thank S. Isakov, D. Ish, K. Najafi, and E. Parker for useful discussions and comments on this manuscript. We would also like to thank the organizers of the workshop  \textit{Theoretical Physics for Machine Learning}, which took place at the Aspen Center for Physics in January 2019, for stimulating this collaboration and project.

\bibliographystyle{JHEP}
\bibliography{refs}

\appendix
\section{Real non-volume preserving flows \label{sec:nvpflow}}
In this appendix we discuss the affine layers which compose real NVP flows in more detail and introduce some notation we found useful.

The efficient invertibility of the flow $G$ derives from the efficient invertibility of the affine layers, which in turn relies on the fact that each layer acts only on a subset of the variables. Denoting this subset as $\mathcal{A}$, and letting $\mathcal{B} = \mathcal{A}^C$ be the complement, then the action of a single affine layer on a variable $y \in \mathbb{R}^N$ is given by:
\begin{subequations}
\begin{equation}
g_a(y; \mathcal{A}, s, t) := t_a \left( y_{\mathcal{A}} \right) + y_a \, \exp\left( s \left( y_{\mathcal{A}} \right) \right)_a  \,,
\end{equation}
\begin{equation}
g_b(y; \mathcal{A}, s, t) := y_b \,,
\end{equation}
\end{subequations}
for $a \in \mathcal{A}$, $b \in \mathcal{B}$. Furthermore, each layer depends on two functions, $s, t: \mathbb{R}^{|\mathcal{A}|} \rightarrow \mathbb{R}^{|\mathcal{B}|}$, which we shall take to be neural networks. The inverse of an affine coupling layer may be easily worked out to be
\begin{subequations}
\begin{equation}
g_a^{-1}(y; \mathcal{A}, s, t) = \left(y_a - t_a \left( y_{\mathcal{A}} \right)\right) \exp\left( -s \left( y_{\mathcal{A}} \right) \right)_a \,,
\end{equation}
\begin{equation}
g_b^{-1}(y; \mathcal{A}, s, t) = y_b \,.
\end{equation}
\end{subequations}
Importantly, note that the functions $s,t$ need not be inverted in order to construct the inverse of the affine layer (in fact, they do not even need to be invertible). The Jacobian matrix $\partial g(y)_i/\partial y_j$ can be shown to have a block-triangular form, and therefore the determinant may be efficiently calculated to be:
\begin{equation}
\det \left( \frac{ g(y; \mathcal{A}, s, t) }{\partial y} \right) = \exp\left( \sum_{a \in \mathcal{A}} s \left( y_a \right) \right) \,.
\end{equation}

Real NVP flows may be constructed by composing multiple affine layers, each with their own $(\mathcal{A}, s, t)$. Denoting the $\ell$-th afffine layer as $g_{(\ell)}$, the flow is defined as
\begin{equation}
    G_{(\ell)} := g_{(\ell)} \circ G_{(\ell-1)} \,,
\end{equation}
for $\ell = 0, 1, ..., L$, with the understanding that $G_{(0)}$ is the identity and $G_{(L)}$ corresponds to the complete flow. The output of each layer of the flow represents a random variable $z_{(\ell)} := G_{(\ell)}(z_{(0)})$, with $z_{(0)} = z$ and $z_{(L)} = x$. The determinant for the full generator map $x = G(z)$ is then simply the product of the determinants for each individual affine layer:
\begin{align}
\det \left( \frac{\partial G(z)}{\partial z} \right) = \prod_{\ell=1}^{L} \exp\left( \sum_{a \in \mathcal{A}^{(\ell)}} s^{(\ell)} \left( z_a^{(\ell-1)} \right) \right) \,.
\end{align}
This equation, used in conjunction with the change of variable formula Eq.~\ref{eq:changeofvar_x}, allows the likelihood of a sample $x$ to be efficiently determined.

\subsection{Implementation details \label{sec:selflearning_implementation}}
The details of the normalizing flow are as follows. For the prior we used an isotropic Gaussian with unit variance. The flow function $G$ was composed of $L=4$ affine coupling layers. For each even layer $\ell = 2k$, the set of variable indices which the flow acts upon non-trivially, i.e. $\mathcal{A}^{(2k)}$, was chosen randomly. Each index was included in $\mathcal{A}^{(2k)}$ with probability 1/2. For each odd layer, the complement was used, $\mathcal{A}^{(2k+1)} = \mathcal{A}^{(2k) C} = \mathcal{B}^{(2k)}$, so that each variable is acted upon by the flow every 2 layers. Each layer used a separate set of neural networks $s^{(\ell)}, t^{(\ell)}$, each with their own set of weights. The neural networks were taken to  be multi-layer perceptrons (MLPs) with 3 hidden layers and one final layer, each with $N$ units.\footnote{As described here, the neural networks are functions from $\mathbb{R}^N$ to $\mathbb{R}^N$, whereas in Appendix \ref{sec:nvpflow} we indicated that $s^{(\ell)}, t^{(\ell)}: \mathbb{R}^{|\mathcal{A}^{(\ell)}|} \rightarrow \mathbb{R}^{|\mathcal{B}^{(\ell)}|}$. This discrepancy is due to the fact that for ease of implementation we considered the full set of $N$ variables and used a binary mask to enforce the fact that the flow should only act upon the set of indices $\mathcal{A}^{(\ell)}$.} Lastly, the activation function for each layer in both networks was taken to be LeakyRelu, except for the final layers. The final activation for $s^{(\ell)}$ was $\tanh$, whereas the final activation for $t^{(\ell)}$ was the identity. 

For the reverse KL-based learning we employed a trick introduced by \cite{li2018neural}, which encouraged the trained flow to approximately respect the $\mathbb{Z}_2$ reflection symmetry of the physical distribution, corresponding to invariance under $x \rightarrow -x$. The trick consists of replacing $p_G(x)$ in the objective function Eq.~\ref{eq:reverseKL} with the symmetrized version ${p_G^{(\text{sym})}(x) = \left( p_G(x) + p_G(-x) \right)/2}$.

In both cases we performed the training by minimizing the corresponding objective with the Adam optimizer \cite{kingma2014adam} with learning rate $10^{-4}$, and used batch sizes of 50. For the reverse KL training, we chose to perform 250,000 mini-batch updates.\footnote{Note that in this case there is no dataset, and so the standard concept of training epochs does not apply.} For the forward KL training we also performed 250,000 updates, which is equivalent to 125 epochs (entire passes through the data) with each epoch consisting of 100,000/50 = 2000 mini-batch updates.
\end{document}